\documentclass[letterpaper]{article} 
\usepackage{aaai24}  
\usepackage{times}  
\usepackage{helvet}  
\usepackage{courier}  
\usepackage[hyphens]{url}  
\usepackage{graphicx} 
\usepackage{natbib}  
\usepackage{caption} 
\usepackage{algorithm}
\usepackage{algorithmic}

\usepackage{newfloat}
\usepackage{listings}
\DeclareCaptionStyle{ruled}{labelfont=normalfont,labelsep=colon,strut=off} 
\floatstyle{ruled}
\newfloat{listing}{tb}{lst}{}
\floatname{listing}{Listing}
\nocopyright

\title{Customized Information and Domain-centric Knowledge Graph Construction with Large Language Models}
\author{
    Frank Wawrzik\equalcontrib\textsuperscript{\rm 1},
    Matthias Plaue\equalcontrib,
    Savan Vekariya\textsuperscript{\rm 1},
    Christoph Grimm\textsuperscript{\rm 1},
}
\affiliations{
    \textsuperscript{\rm 1}University of Kaiserslautern-Landau (RPTU)\\

    Gottlieb Daimler Straße\\
    67663 Kaiserslautern, Germany\\
    wawrzik@cs.uni-kl.de, plaue@mapegy.com, svekariy@rptu.de, grimm@cs.uni-kl.de
}

\usepackage{bibentry}

\begin{document}

\maketitle

\begin{abstract}
In this paper we propose a novel approach based on knowledge graphs to provide timely access to structured information, to enable actionable technology intelligence, and improve cyber-physical systems planning.
Our framework encompasses a text mining process, which includes information retrieval, keyphrase extraction, semantic network creation, and topic map visualization. Following this data exploration process, we employ a selective knowledge graph construction (KGC) approach supported by an electronics and innovation ontology-backed pipeline for multi-objective decision-making with a focus on cyber-physical systems. We apply our methodology to the domain of automotive electrical systems to demonstrate the approach, which is scalable.
Our results demonstrate that our construction process outperforms GraphGPT as well as our bi-LSTM and transformer REBEL with a pre-defined dataset by several times in terms of class recognition, relationship construction and correct "sublass of" categorization. Additionally, we outline reasoning applications and provide a comparison with Wikidata to show the differences and advantages of the approach.
\end{abstract}

\section{Introduction}

The planning of modern cyber-physical systems necessitates the ability to predict technology trends, employ collaborative community approaches, allocate tasks efficiently, develop design roadmaps, and even facilitate self-organizing systems. In the era of digitalization, these processes demand quicker decision-making, often amidst increasingly competing interests and considerations.
Technology Intelligence is a data analytical approach to evaluate and follow a technologies potential and behaviour.
Making these decisions relies on the availability of timely and pertinent information from extensive datasets. One effective strategy for managing this complexity is the integration of Large Language Models (LLMs) into one's document database and utilizing specific questions or prompts. However, it is worth noting that these approaches are currently not machine-actionable. Therefore, in this paper, we propose a knowledge graph-based method aimed at enhancing technology intelligence and facilitating machine-actionable innovation construction to improve the planning of cyber-physical systems.

Conventional methods for constructing knowledge bases typically concentrate on a specific domain and rely on a dedicated corpus. This corpus is often relatively static or updates predefined properties outlined in ontologies using dedicated datasets. Frequently, only a limited number of distinct sources are crawled and incorporated into the knowledge graph, such as Wikipedia articles or research articles from a particular journal. The challenge here is that these knowledge graphs often fall short in terms of succinctness and completeness, both of which are critical quality criteria.

In this paper, we introduce a novel approach that involves the pre-classification of information from heterogeneous sources for constructing the knowledge graph. Our methodology involves crawling a wide variety of sources at an extensive scale. This strategy mitigates the drawbacks associated with attempting to crawl the entire internet and constructing an overly voluminous and irrelevant graph. Simultaneously, it results in the accumulation of more relevant information, contributing to the creation of larger and more comprehensive knowledge bases. Additionally, we complement this approach with reasoning techniques to enhance semantic accuracy.

The contribution of the paper is a two stage knowledge graph construction process. By preselecting only relevant documents for the domain the KG construction via large language models, it can construct tailored domain knowledge.

\begin{figure*}[bt]
\centering
\includegraphics[width=0.95\textwidth]{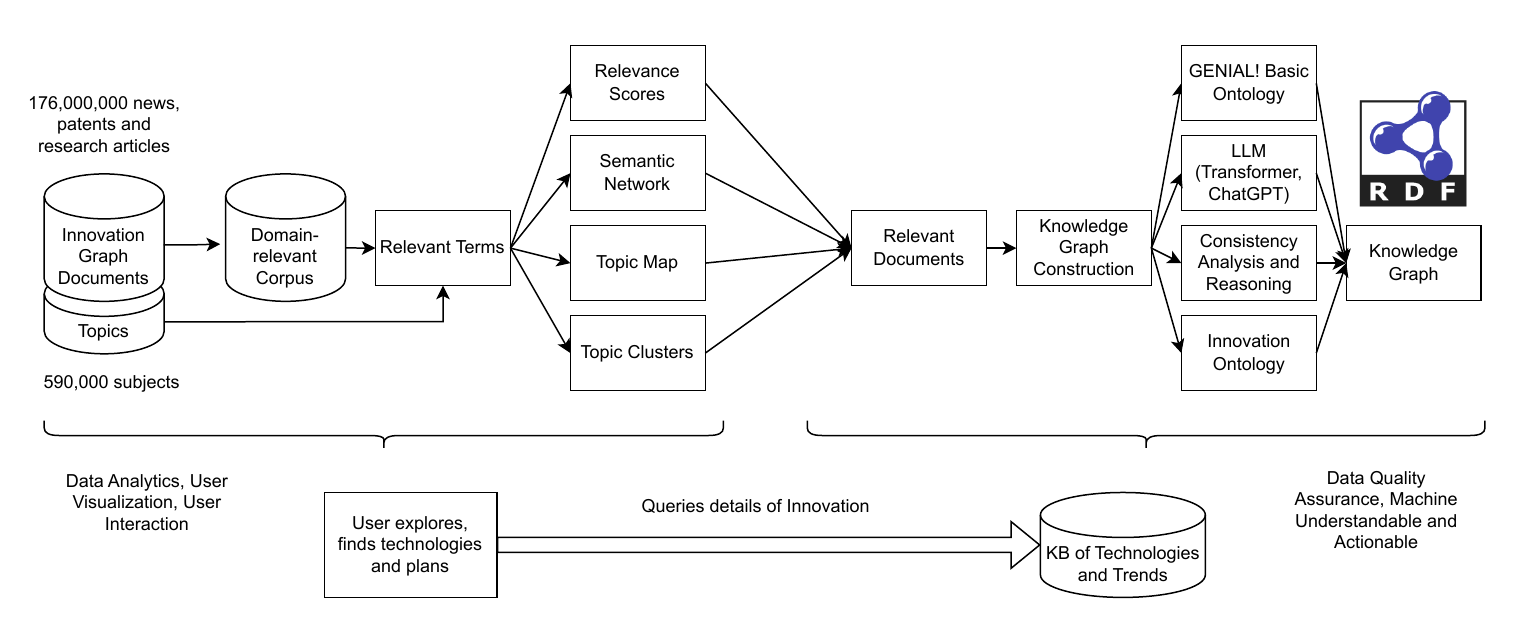}
\caption{Pre-refined knowledge graph construction framework for innovation planning}
\label{fig:scouting_framework}
\end{figure*}

\section{State of the Art}

\subsection{Text mining for technology intelligence}


Technology intelligence ``provides an organisation with the capability to capture and deliver information in order to develop an awareness of technology threats and opportunities'' \cite{Kerr2006}. One important subfield of technology intelligence is technology mining, or tech mining for short: ``the application of text mining tools to science and technology information, informed by understanding of technological innovation processes'' \cite[p.19]{Porter2004}.

Methods from natural language processing (NLP) can help technology scouts explore a large collection of documents, such as news articles, more efficiently by detecting events and trends \cite{Panagiotou2021}, or summarizing key points \cite{Ma2022}. Removing irrelevant results such as fake news  \cite{Capuano2023} can also reduce the data deluge.

\subsection{Topic maps and semantic networks}

One important tool in tech mining are topic maps: visual representations of relations among terms (e.g., technologies, organizations) or among documents. Topic maps have been proposed for many years as an essential tool for patent landscaping and scientific domain analysis \cite{MoyaAnegn2004, Yang2010, Kay2014, Hofmann2019}. Topic clusters are related terms that belong to a topic.

Topic maps can be implemented by visualizing semantic networks based on semantic similarity \cite{Sarica2020, Kim2023} using graph drawing techniques. With the proliferation of increasingly powerful LLMs that allow for the effective semantic embedding of text, these methods present an attractive alternative to more traditional metrics like co-citation or classification overlap. According to \cite{Muennighoff2022}, the performance of text embeddings on different tasks ``varies strongly with no model claiming state-of-the-art on all tasks.'' However, language models adapted to a specific domain can produce embeddings that allow for high performance on downstream NLP tasks applied to data from that domain---for example, analyzing scientific publications \cite{Beltagy2019, Cohan2020, Singh2022}.



\subsection{Knowledge graph construction}
Current knowledge graph construction methods utilize forms of machine learning for example with deep learning or transformers \cite{https://doi.org/10.48550/arxiv.1909.03193,huguet-cabot-navigli-2021-rebel-relation,10.1007/978-3-031-20627-6_24}. However utilizing LLM's for knowledge graph construction is new.
In  \cite{10.1007/978-3-031-47240-4_22} the authors investigate if LLMs can be effectively used to generate ontologies and knowledge graphs via the zero-shot prompting method. They use lexical terms, taxonomic discovery and triple relations. Whereas their work was to evaluate the general suitability, we constructed a knowledge graph based on definitions in a reference ontology. \cite{LOPES2023110385} classify domain entities by allocating those entities to their corresponding top level superclass. They use their own trained transformer-based language models. GraphGPT \cite{tang2023graphgpt} has to be mentioned as related work. It is able to generate more detailed graphs than transformer models, but lacks structure and quality criteria which is both improved within this work.

\subsection{Planning and electronics ontologies}
\cite{taskontology} introduces a task ontology for domain independent planning. \cite{10.1007/978-3-319-70833-1_35} also proposes an task-ontology-based approach to improve coordinated planning and problem solving of autonomous agents. Other works focus role of ontologies in CPS \cite{rolecps} or general surveys in this area to investigate automatic construction \cite{kgmlcps}. Only this work however generates large knowledge graphs, which can be used to identify developments and serve as a general knowledge base for planning for CPS.\\
A comprehensive overview of the foundations of knowledge engineering for cyber-physical systems can be found in \cite{Wawrzik2022}. The GENIAL! Basic Ontology, based on an upper ontology and the ISO26262 standard on automotive safety, describes cyber-physical systems and digital twins in detail \cite{DBLP:conf/semweb/WawrzikL21}. 

\section{Method}

\subsection{Framework for technology intelligence andpPlanning - overview and approach}
Figure~\ref{fig:scouting_framework} visualizes the framework and its process in detail.
First, we query a technology intelligence database for relevant documents such as research articles and patents. From this corpus, we extract keyphrases that help diversify, spread, and improve search results. We also use the keyphrases to create a semantic network, and visualize this network as a topic map. The topic map helps users interactively explore the domain. Based on these methodologies we arrive at relevance scores for each desired technology, product, trend, or innovation. This information allows the user to select the documents with highest relevance and importance to the field and use case at hand, that serve as the input for final knowledge graph construction (KGC). In the KGC pipeline, the first step is to convert the articles into .txt files that Owlready2 \cite{Lamy2017} transforms into an OWL Ontology Graph (.owl file). The transformation occurs via the transformer REBEL \cite{huguet-cabot-navigli-2021-rebel-relation}, and large language models like ChatGPT. The REBEL transformer model was pre-trained with an electronic dataset from Wikipedia articles, whereas ChatGPT was optimized with several prompting approaches and functions. Additionally, the created knowledge graph is applied to a reasoning procedure that commits to definitions that are defined in the GENIAL! Basic Ontology as the knowledge graphs schema. This reasoning procedure ensures an improved consistency and structure, and improves the results of the machine learning as well. GBO is capable of describing all hardware and software systems - from smart homes and refrigerators to autonomous cars, cyber-physical energy systems and graphics processing units. This approach is generalizable to any domain in as much as the prompt is adapted to a new vocabulary of the relevant terms of the domain. Further the content of articles needs to reflect the relevance for the vocabulary, but can show heterogenity and diversity. To adapt the pipeline to a new knowledge source then just requires: 1) adding textdocuments of the new domain  and 2) defining keywords for search. Additionally important to mention that currently a knowledge expert is still required at the last stage to supervise the quality of the content. 

\subsection{Innovation Graph database}
As of November 2023, the our Innovation Graph database includes 50 million research publications, 81 million internet news articles, and 45 million patent filings from the past 10 years. We will refer to documents from these different domains as document \textit{genres}, in order to avoid confusion with the knowledge domain of innovation and technology.

The titles and abstracts (when available) of these documents have been matched against a thesaurus that includes 590,000 terms relevant to global innovation and technology.

\subsection{Information retrieval and keyphrase extraction}
Tech mining distinguishes itself from ``data mining and text mining by its reliance on science and technology domain knowledge to inform its practice'' \cite[p.19]{Porter2004}. In practice, domain knowledge is injected into the text mining process through the definition of a search strategy that is used to extract the relevant documents under study from the database \cite{MacFarlane2022}.

Agreeing with \cite{Comai2018}, we argue that in order to cover the whole technology ecosystem, that this corpus of relevant documents be drawn from three genres: research publications, internet news articles, and patent documents.

Once the corpus has been extracted, we can identify the most relevant terms from the thesaurus by co-occurrence analysis. More specifically, for each genre and term, we compute the normalized pointwise mutual information (nPMI), and keep those terms that are associated with a positive value for every genre.

\subsection{Semantic network}
Next, we build a semantic network --- the nodes of which are defined by the most relevant terms extracted in the previous step. First, we need to embed the terms, so we need to select a model for text representation. Benchmarks that compare the current state-of-the-art in terms of accuracy are readily available. 


In order for the embedding to better capture the meaning of each term, we compute the embedding of the term concatenated with a short description which, in the majority of cases, had been extracted from Wikipedia.

We tested two methods to construct the final semantic network from these embeddings:
\begin{itemize}
\item simple thresholding, connecting any two terms the embeddings of which have cosine distance of at most 0.5,
\item a technique called ``semantic co-occurrence'' where each topic is represented by their cosine distance with every document in corpus, which can be interpreted as a fuzzy membership function---similarity of two topics can be defined by the Tanimoto similarity of those memberships.
\end{itemize}


\subsection{Ontology for Innovation and Planning}
\begin{figure*}
\centering
\includegraphics[width=0.9\textwidth]{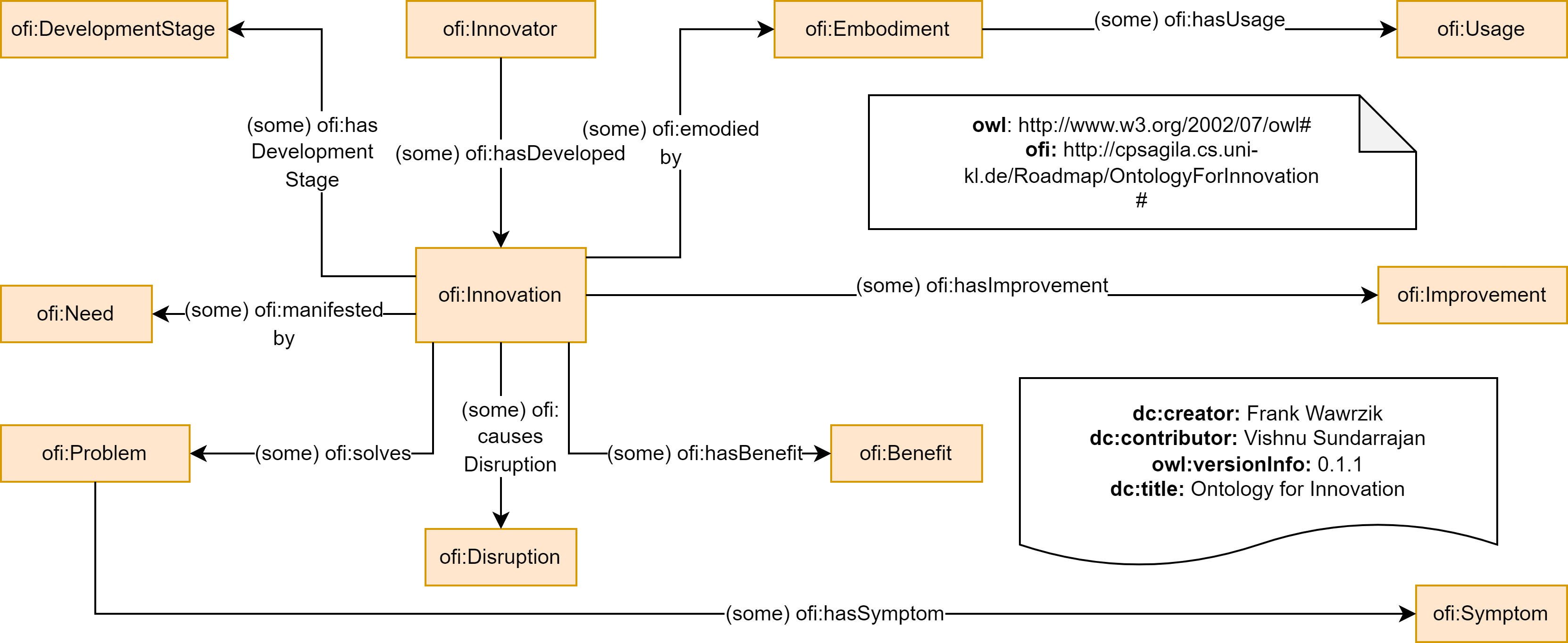}
\caption{Ontology for innovation planning}
\label{fig:innovation}
\end{figure*}

The Ontology for Innovation\footnote{http://www.lexicater.co.uk/vocabularies/innovation/ns.html} was built in 2012 by an innovation center and Volkswagen. It represents well a divers and informal domain in a way where complex interdependencies can be discovered and expressed. Figure \ref{fig:innovation} shows a rebuilt domain model based on the chowlk notation\footnote{https://chowlk.linkeddata.es/notation.html}. For example innovations usually satisfy a need that accounted for the innovation. Every innovation has a stage from defining the need to distributing the product. Further they are related to an improvement of for example efficiency or quality of something. Also innovations are assigned to disruptions. For example quantum computing disrupts cryptography. Additionally, each innovation is allocated to a design state from conceptualization to distribution. A populated knowledge graph based on the innovation ontology is thus able to predict developments and bring disruptions to planning either to the awareness of the user or to act on it autonomously.

\begin{figure*}[bt]
\centering
\includegraphics[width=0.99\textwidth]{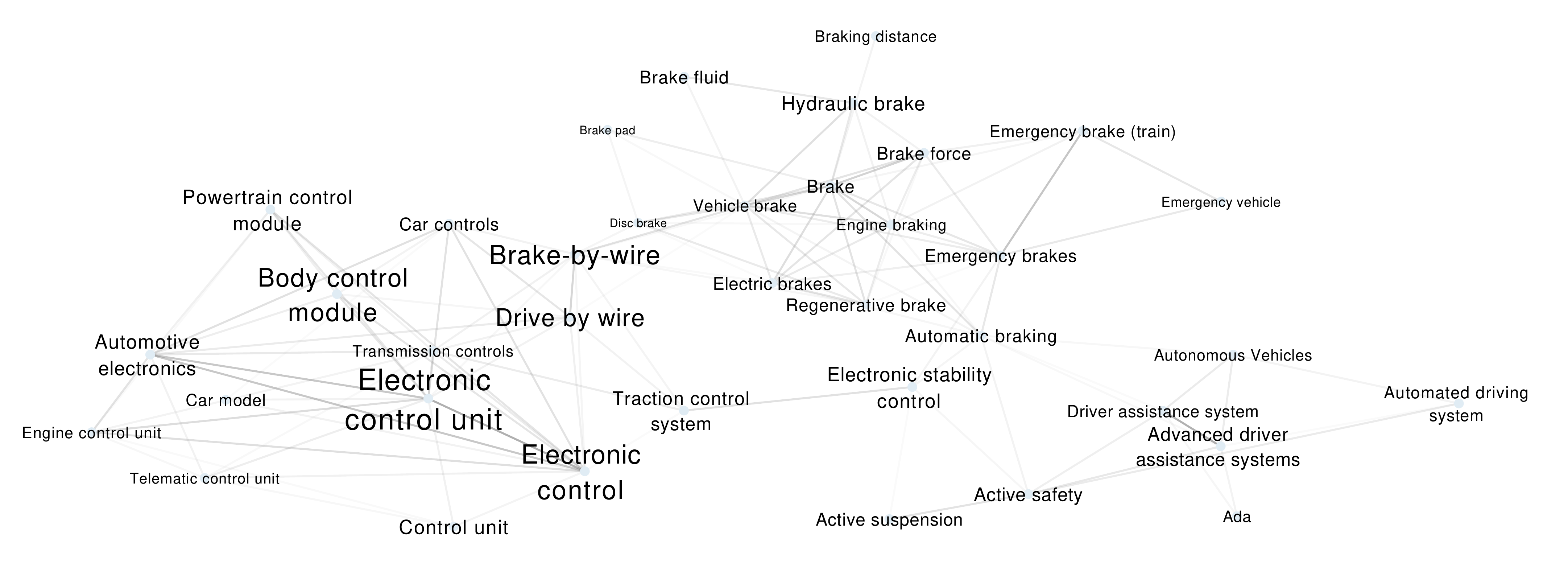}
\caption{Example of a cluster within our semantic network ``automotive electrical systems''}
\label{fig:example_cluster}
\end{figure*}


\section{Evaluation}
\subsection{Use Case: Automotive Electrical Systems Motivation}
Automotive electrical systems are intricate networks that perform several vital functions, including power distribution, data transmission, and control. These systems are characterized by a wide array of components, such as Electronic Control Modules (ECMs), sensors, and cables. Additionally, they integrate advanced technologies like Advanced Driver Assistance Systems (ADAS), Ethernet communication, and Over-The-Air (OTA) capabilities. 

Given their complexity, automotive electrical systems represent an excellent example of the benefits of employing knowledge graphs. These graphs facilitate the mapping of complex dependencies and interactions among various components and technologies. This mapping provides a comprehensive overview crucial for effective decision-making and planning. For example, when introducing a new component, a knowledge graph can demonstrate how this addition will interact with existing systems, predicting potential conflicts or synergies.
\subsection{LLMs to create semantic networks for data analytics}
As stated in the semantic network section in the methods section, we also used large language models to create our semantic network of keywords and also chose different LLMs.
However, tf--idf (term frequency--inverse document frequency) and fastText \cite{grave2018learning} are much faster on inference (see Table~\ref{tab:benchmark_runtime}) which motivated us to perform our own comparison on some simple downstream classification tasks, see Table~\ref{tab:benchmark_accuracy}.
The more recent language models that we compared to ``classic'' approaches were the large, English BGE model \cite{bge_embedding} and the lightweight all-MiniLM-L6-v2 model derived from \cite{wang2020minilm}.
The tasks that we based this comparison on were for a $K$-nearest neighbor (cosine distance) classifier to determine the classes defined by a lexical search strategy (\#1) or a taxonomy Sectors (\#2), respectively, processing the documents' (a) titles and abstracts, or (b) only the titles. Based on this evaluation, we decided to use the all-MiniLM-L6-v2 model which shows the best trade-off between runtime performance and accuracy.
\begin{table}
    \centering
    \begin{tabular}{l|c|c|c|c}
\textbf{model}&\textbf{size}&\textbf{GPU}&\textbf{CPU}&\textbf{quant.}\\
\hline
BGE large&	1.3 GB&	33&	3&	7\\
MiniLM L6&90 MB&113&	47&	58\\
fastText&	4.2 GB&	-&	2000+&	-\\
tf--idf&	vocab.&	-&	400+&	-
    \end{tabular}
    \caption{Runtime performance of inference with different embeddings in data records per second}
    \label{tab:benchmark_runtime}
\end{table}

\begin{table}
    \centering
    \begin{tabular}{l|c|c|c|c|c}
\textbf{model}&\textbf{year}&\textbf{\#1a}&\textbf{\#2a}&\textbf{\#1b}&\textbf{\#2b}\\
\hline
BGE large&	2023&	\textbf{82\%}&	\textbf{63\%}&	\textbf{78\%}&	\textbf{60\%}\\
MiniLM L6&2020&80\%&	\textbf{63\%}&	77\%&	\textbf{60\%}\\
fastText&2018&	71\%&	56\%&	68\%&	54\%\\
tf--idf	&$<$ 1990&73\%&	56\%&	70\%&	54\%
    \end{tabular}
    \caption{Accuracy achieved on downstream tasks for different embeddings}
    \label{tab:benchmark_accuracy}
\end{table}

\subsection{Topic map visualization for automotive electrical systems}
In order to extract the corpus from the Innovation Graph database, the search strategy in Listing~\ref{list:search_sstr} was used. 
Search terms have automatically been stemmed by the system, and the publication dates (filing dates for patents) have been restricted to the range between 2018-10-13 and 2023-10-12.

As could be expected, many of the top relevant extracted thesaurus terms with highest nPMI are included with the domain expert's original search strategy. Table~\ref{tab:nPMI} shows the top 10 terms that were newly detected, and not part of the original search strategy.

The semantic network for ``automotive electrical systems'' consists of 708 nodes and 2836 edges. We used the network analysis software Cytoscape \cite{shannon2003cytoscape} to draw and layout the network. In addition, the Markov Cluster Algorithm (MCL) was used to cluster the network \cite{morris2011clustermaker, VanDongen2008}. This process resulted in the generation of 120 clusters, with 60 clusters being comprised of more than two nodes. One example cluster is illustrated in Figure~\ref{fig:example_cluster}.

\begin{figure}[h]
\centering
\includegraphics[width=0.99\columnwidth]{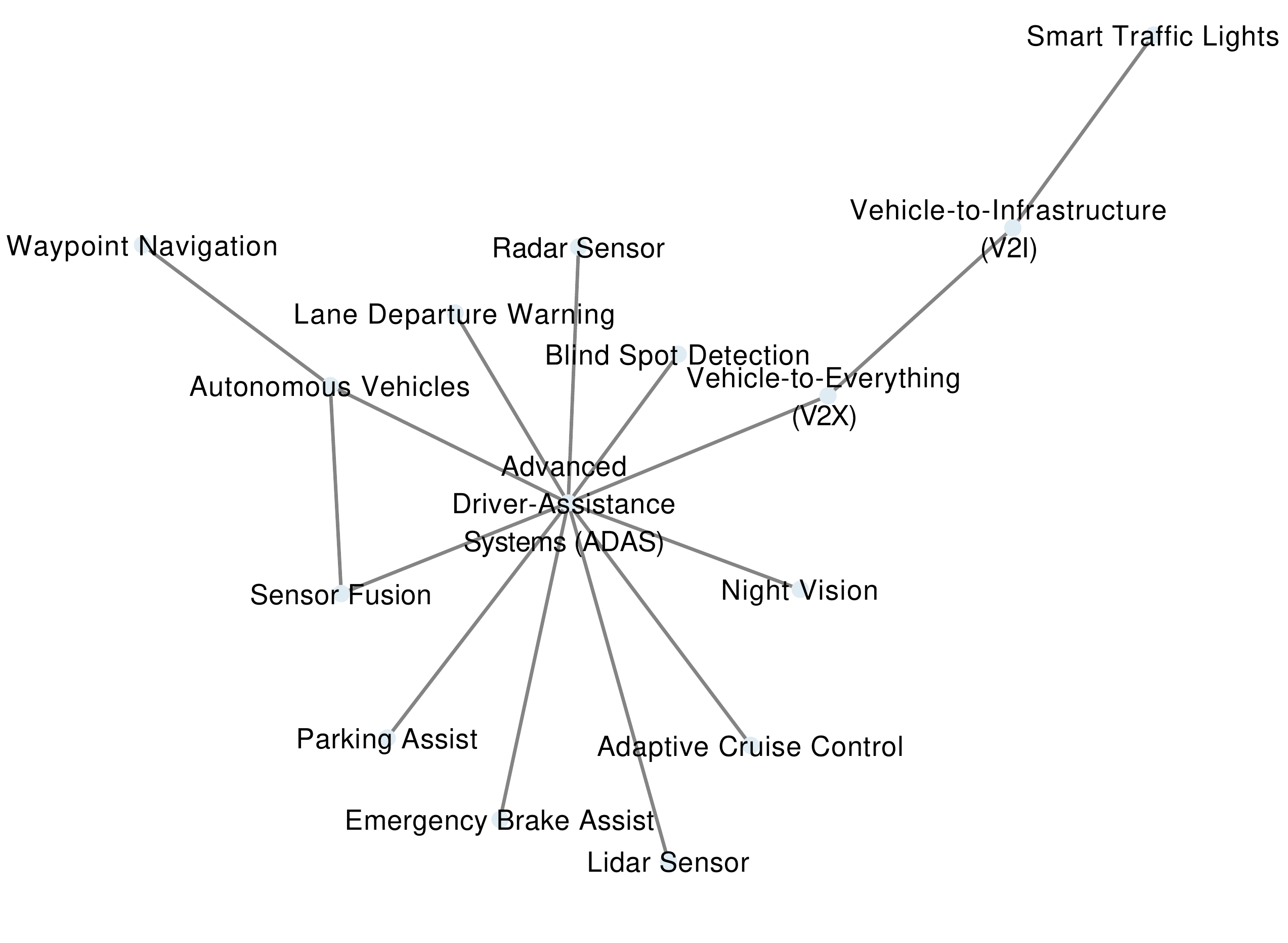}
\caption{Example cluster within a semantic network generated by prompting ChatGPT}
\label{fig:example_chatgpt}
\end{figure}

In order to present an alternative to our approach, we prompted ChatGPT to produce an edge list of related terms in the field of ``automotive electrical systems'' \cite{GPTChat}. Figure~\ref{fig:example_chatgpt} shows part of that semantic network.

\begin{listing}
(``automotive'' OR ``automobile'' OR ``car'' OR ``truck'' OR ``bus'') AND (``wiring system'' OR ``board network'' OR ``e/e architecture'' OR ``electrical/electronic architecture'' OR ``centralized architecture'' OR ``smart architecture'' OR ``cross-domain'' OR ``domain-oriented'' OR ``zonal architecture'' OR ``domain-centric'' OR ``zone controller'' OR ``eea'' OR ``electrical system'' OR ``electrical infrastructure'' OR ``intra-vehicle network'' OR ``controller area network'' OR ``communication circuit'' OR ``digital transmission'' OR ``data transmission'' OR ``data transfer'' OR ``next-generation connectivity'' OR ``control module'' OR ``domain controller'' OR ``sensor interface'' OR ``vehicle-to-vehicle'' OR ``vehicle-to-grid'' OR ``vehicle-to-device'' OR ``electronic control unit'' OR ``powertrain control module'' OR ``power-train control module'' OR ``power electronics'' OR ``power semiconductor'' OR ``drive-by-wire'' OR ``drive by wire'')
\caption{Search strategy for matching documents relevant to ``automotive electrical systems''}
\label{list:search_sstr}
\end{listing}

\begin{table}
\begin{tabular}{l|c|c|c}
\textbf{term}&\textbf{news}&\textbf{science}&\textbf{patents}\\
\hline
Brake-by-wire&0.594&0.565&0.502\\
Body control module&0.686&	0.547&	0.486\\
Vehicle electrics	&0.384&	0.497&	0.381\\
Local Interconnect Network&	0.523&	0.595&	0.371\\
Vehicle bus	&0.366&	0.468&	0.358\\
Sensor interfaces	&0.440&	0.419&	0.353\\
Data bus	&0.350&0.457	&0.394\\
Bus communication	&0.343	&0.535&	0.448\\
Bus (computing)&0.330&0.499&0.363\\
Vehicle electronics&0.417&0.512&0.328
\end{tabular}
\caption{Top 10 newly detected relevant terms according to minimum normalized PMI across genres}
\label{tab:nPMI}
\end{table}

\subsection{Knowledge graph construction (KGC) for automotive electrical systems}
Using the results of the text mining procedure explained in the previous section, all relevant documents have been identified and are fed to the KGC pipeline. The articles contain descriptions of specific technical contributions related to the field of automotive electrical systems. The knowledge graphs constructed from these documents describe these components in more detail than the topic map, and provide a machine readable overview of the content.

Our previous implementation with a transformer and a bi-LSTM required training several assistants to perform manual tagging on our datasets (see \cite{info14030176}). This was a time consuming task that took around three months with 20 person hours per week. Besides the amount of time spent, it was a challenge to train the assistants. The classification according to the distinctions of GBO were fine-grained and challenging to comprehend for non-ontology and non-domain experts.
The implementation using ChatGPT resulted in several improvements when compared to our previous bi-LSTM with transformer model, known as REBEL. The bi-LSTM achieved an f1 score between 0.36 and 0.78 depending on the amount of tags, the tagging class and the complexity of word recognition (and accuracy of human tagging). We observed an increase in the number of recognized classes and the construction of relationships as well. In contrast to REBEL, ChatGPT generated a greater number of classes and offered more descriptive labels. The increase of number of classes was significant with just over 200\% for the LLM for our three article dataset. The amount of correctly classified classes increased as well.

The prompting had to be carefully crafted and optimized in order to improve classification accuracy according to the GENIAL! Basic Ontology (GBO). We taught ChatGPT our vocabulary and relationships, experimented with wording and the stages of prompt design (context, persona, task, format, exemplar, and tone). 

Due to the large amount of data in the corpus, we selected three research articles with a high relevance score to demonstrate the approach:
\begin{itemize}
\item Optimal Operation of Automotive Electrical
System with Photovoltaic Generation and Three-level Battery Management Scheme 
\item Designing Attacks Against Automotive Control Area Network Bus and Electronic Control Units 
\item CAN-FT: A fuzz testing method for automotive
controller area network bus 
\end{itemize}

\begin{figure}[bt]
\centering
\includegraphics[width=0.32\textwidth]{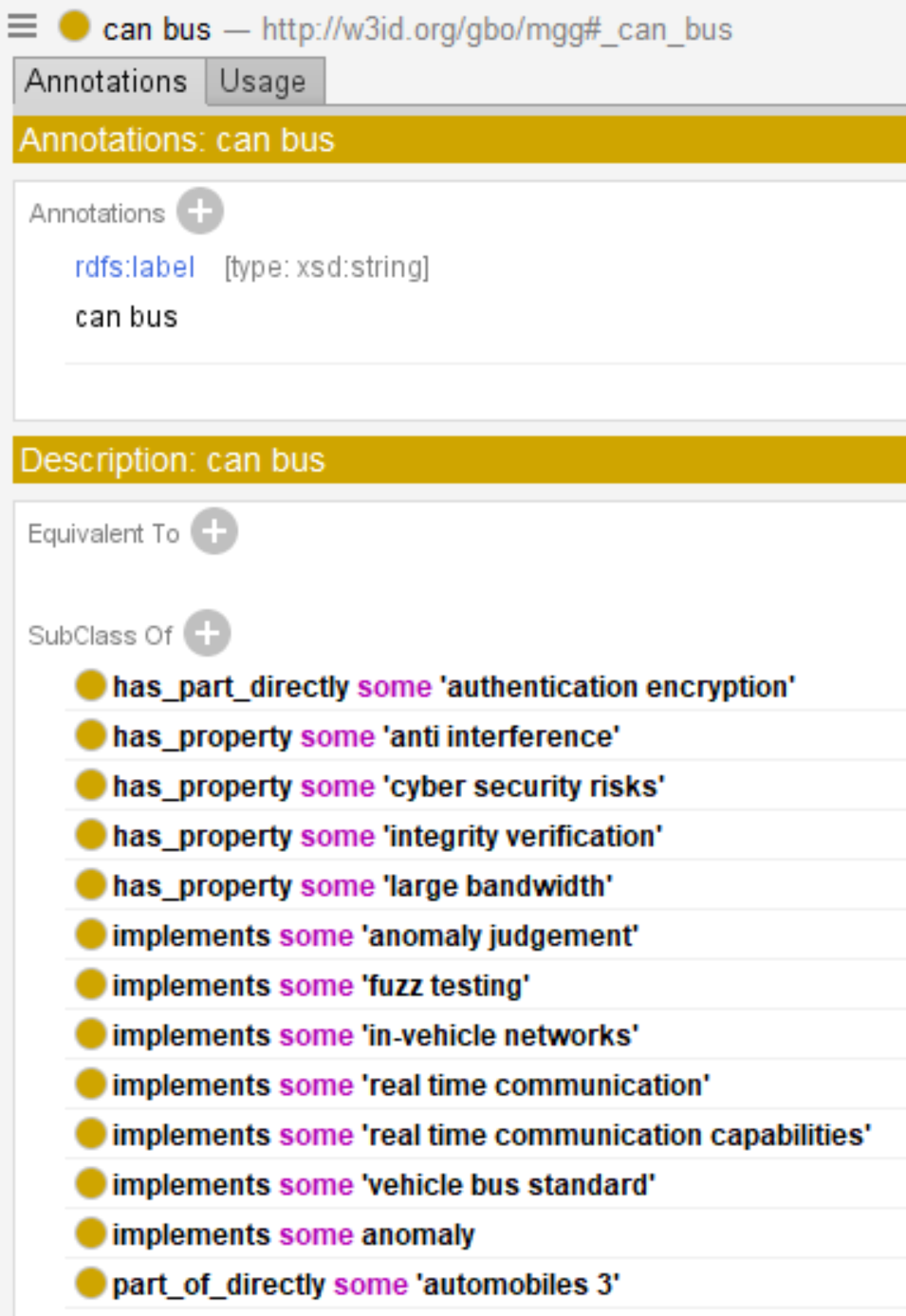}
\caption{Relationships created for class CAN-Bus}
\label{fig:CANBus}
\end{figure}

\begin{figure*}[ht]
\centering
\includegraphics[width=0.9\textwidth]{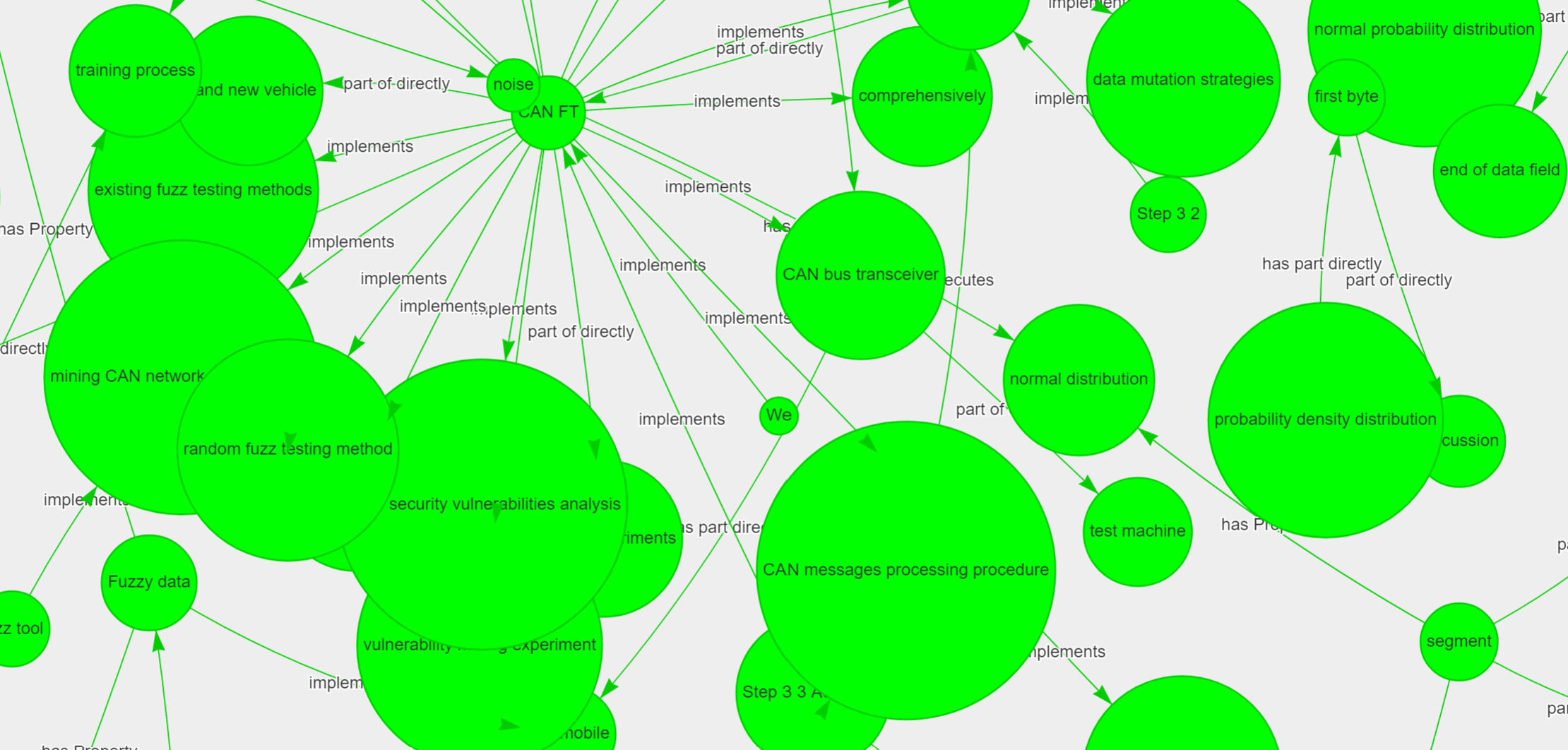}
\caption{Graph view of CAN-FT triples}
\label{fig:CANft}
\end{figure*}

As illustrated in Figure \ref{fig:CANft}, the articles are related to automotive electrical systems, but cover a wider range of issues. In the Figure we see a graph in the pickle format visualized in html. In comparison to prot\'{e}g\'{e}, graph triples can be identified easily. The CAN FT bus and nearby nodes are shown. The "implements" relationship links to functions, and the "part of directly" relationship to the next connected upper hierarchy. Functions that the CAN FT can supply are for example existing fuzz testing methods, messaging processing procedures or security vulnerability analysis. We see a variety of classes and nodes ranging from cyberattacks to testing methods and new technology compositions.
The constructed knowledge graph contains 3100 axioms, 650 classes and 16 object properties. The low object property count increases reusability and reduces complexity, and is an intentional design decision. Further labels were annotated for natural language spaces and synonym recognition, which also is a best practice in the semantic web. Figure~\ref{fig:CANBus} shows the class ``CAN-Bus'' and the relationships (existential restrictions) constructed for this class. Upon examining the extracted properties and comparing them with the content of the research articles, we discovered that content from all three papers has been included in the constructed CAN-Bus class. We showcase our results with three papers, because the results are representative of the generated content as a whole and would not be manageable to human examination otherwise. The scaled graph compares closely. 

\subsection{Reasoning application}
\begin{figure}[H]
\centering
\includegraphics[width=0.32\textwidth]{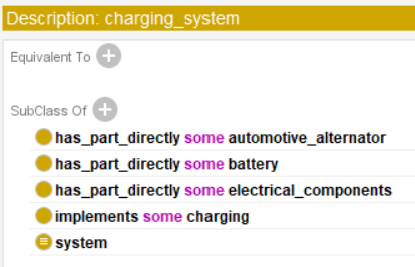}
\caption{Charging System passing the consistency checking}
\label{fig:reasoning}
\end{figure}
After the construction process our graph is checked against the definitions of our reference ontology GBO, which was introduced earlier. This is work in progress and a challenge due to two reasons: long reasoning times and the complexity of expressiveness as well as the amount of data to be supervised. Figure \ref{fig:reasoning} illustrates the idea and gives an example. The Figure shows the class charging system, which was correctly classified as a system. Further the relationships (existential restrictions) of the charging system are shown. Both the automotive alternator and the battery were classified as hardware components and the electrical components as components. The difference here is that components are allowed to contain software. Hardware components are a subclass of components that constitute of hardware parts. As these components are in the next hierarchical level the reasoner test is passed. Charging was classified as function, and thus the range axiom of the implements relationship is also fulfilled. This makes the charging system a valid entry in our knowledge base and is thus not removed.

\subsection{Comparison with Wikidata}
Wikidata \cite{Wikidata2014} is one of the largest structured knowledge bases. It contains classes and relationships of almost any domain. We examined Wikidata's knowledge base in regard to the domain covered in this work. It has a large upper level to structure its classification, often by multiple inheritance. The relationships where often similar with minor differences in modelling choices in comparison with the GENIAL! Basic Ontology.

For example, our ontology is built for specific reasoning operations. Thus, we have the has\_part\_directly relationship (non-transitive, sub-object property of has\_part) in addition to the has\_part relationship which is transitive. Wikidata uses has\_part as well. This supports keeping our definitions in GBO consistent. Wikidata is human constructed and based on Wikipedia's knowledge sources. We executed tailored queries for our domain and found that there is comparatively few knowledge content. In the area of processors we found various specific processors, but the list was small and exemplary rather than comprehensive and complete. Similar for the more abstract electrical components in general. Our approach with applying it to just three articles already outperformed the Wikidata's size by many times and this by considering a much less known and less represented domain of the automotive electrical system. Thus achieving a higher degree of expert domain knowledge and a significantly higher degree of relationship representations. It is of importance to note that this approach scales well and is easily verifiable via reasoning.

\begin{figure}
\centering
\includegraphics[width=0.44\textwidth]{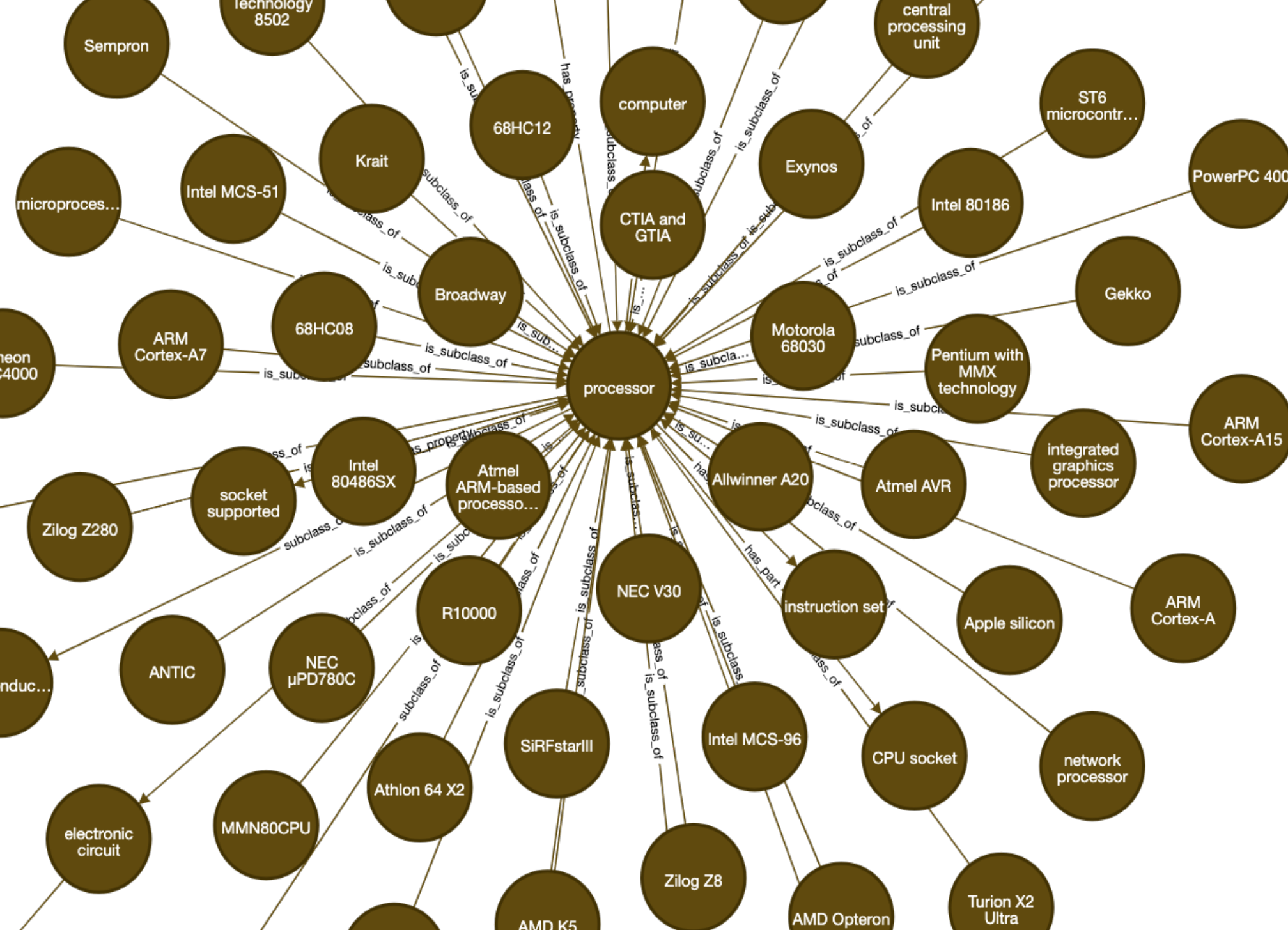}
\caption{Processor class and its relationships (Wikidata)}
\label{fig:processor}
\end{figure}

Figure~\ref{fig:processor} exemplifies the findings of processor. We find mostly processor types, few sublasses and triples comprising about 30 classes. Sensors, actuators, busses, wires, communication networks, amplifiers or especially any other more specific technologies are only marginally covered or not at all.

\subsection{Results of generated Innovation Knowledge Graph}
In our approach, generating knowledge graph TBox triples from text using GPT models is divided into two phases. The first phase involves extracting meaningful relations from a single sentence, focusing on the relationships between entities as shown in Figure~\ref{fig:innovation}. The second phase is about determining the class of entities within the triples. 

For both phases, we have engineered specific prompts that can be fed to the GPT-4 model. This method allows for the initial extraction of triples, followed by the assignment of an appropriate class to these subject-predicate-object relationships. The prompts we used are as follows (excerpt):
\begin{scriptsize}
\begin{verbatim}
As a knowledge graph expert, your task is to extract all 
possible meaningful triples from a given sentence following
specific schema. The schema defines triplets in the format 
{head : ENTITY 1, relation : RELATIONSHIP, tail : ENTITY 2}.
The RELATIONSHIP signifies the relationship between entities.
Relation should be one of the following.
Class I -> Relation -> Class II 
innovator -> has developed -> innovation
innovation -> has development stage -> development stage
innovation -> fulfills -> need
innovation -> solves -> problem
problem -> has symptom -> symptom
innovation -> causes disruption -> disruption
innovation -> has benefit -> benefits
innovation -> has improvement -> improvement
innovation -> embodied by -> embodiment
embodiment -> has usage -> usage
\end{verbatim}
\end{scriptsize}

In these prompts, we specify the schema format, emphasizing the roles of entities and relationships. We also provide examples of how relationships are defined between two classes of entities, such as ``innovator $\rightarrow$ has developed $\rightarrow$ innovation''. 

\begin{figure}
\centering
\includegraphics[width=0.85\columnwidth]{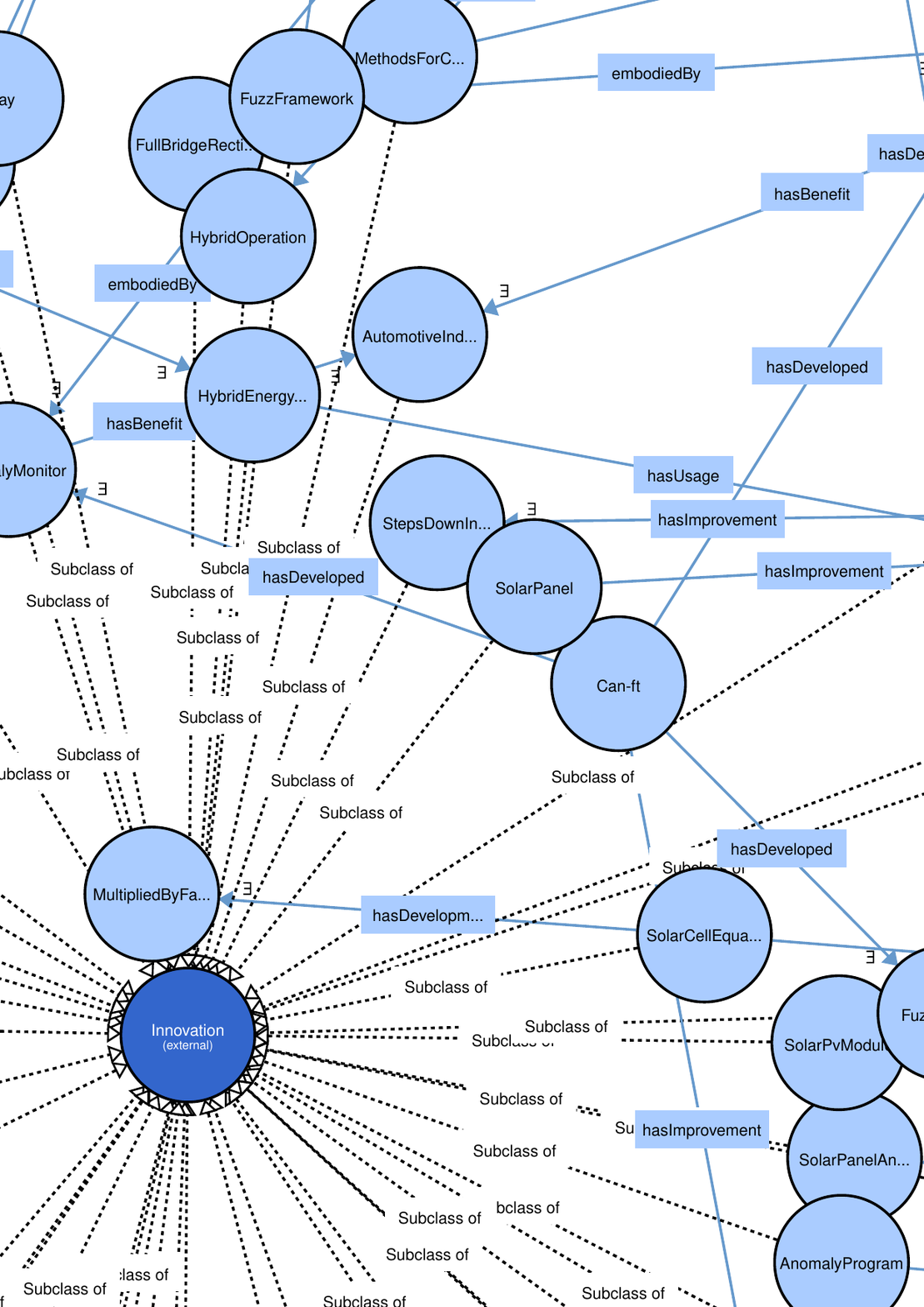}
\caption{Excerpt of generated Innovation Ontology graph}
\label{fig:owl}
\end{figure}

These are some of the triples generated by the GPT-4 model using the aforementioned prompts:
\begin{scriptsize}
\begin{verbatim}
[{"head":"Automotive Alternator","relation":"embodied by",
"tail":"engine"},
{"head":"random fuzzy data","relation":"has symptom",
"tail":"data explosion"},
{"head":"CAN bus","relation":"has benefit",
"tail":"anti-interference"}]
\end{verbatim}
\end{scriptsize}
Figure~\ref{fig:owl} shows a graphical representation of the innovation knowledge graph created triples, where nodes represent entities and edges signify relationships.



\section{Conclusion}

In this paper, we have introduced an innovative knowledge graph construction process that focuses on generating customized information without being confined to a predefined corpus or specific knowledge source. Restricted to three papers to illustrate the framework, we showed a scalable approach, where reasoning time is not a bottleneck. In this way huge knowledge graphs can be constructed automatically. In comparison to alternative approaches, the output of our integrated framework is versatile, catering to both human interpretation and machine processing, making it a valuable resource for technology planners.
We demonstrated that our approach can create high-quality topic maps and knowledge graphs that represent various areas of technology and innovation, such as automotive electronics. These visualizations and structures can assist in complex decision-making for the planning of cyber-physical systems. 
As part of our future endeavors, we aim to enhance the quality assurance aspects of the knowledge graph construction process, ensuring the accuracy of all relationships and classifications. Here correct subclassing of related classes is a challenge (taxonomic correctness), completeness of the knowledge generated.




\section{Acknowledgments}
AI assistant software was employed exclusively for language editing purposes during the preparation of this article. No content produced by text-generating AI, aside from the referenced material, is incorporated into this article.
This article has been funded by the Federal Ministry of Education and Research (Germany), research project KI4BoardNET under grant no. 16ME0782.

\section{Resources}
The generated graph of the example research articles can be found on GitHub\footnote{https://github.com/savanvekariya/Knowledge-Graph-Construction-for-Technology-Intelligence-and-Planning-of-Cyber-Physical-Systems}.

The following data are provided with this repository as well:
\begin{itemize}
    \item Corpus, extracted keyphrases, semantic network,
    \item generated knowledge graph in OWL and HTML formats, generated with ChatGPT
    \item Ontology for Innovation Knowledge Graph in OWL and HTML as outlined in figure \ref{fig:owl},
    \item Wikidata queries and query results in JSON for electronic domain,
    \item links for utilized dataset articles.
\end{itemize}

\bibliography{aaai24}

\end{document}